\documentclass[runningheads]{llncs}
\usepackage{graphicx}
\graphicspath{fig/}
\usepackage{subfig}
\usepackage{booktabs,amsmath}
\usepackage{color}

\begin{document}

\title{Loss Guided Activation for Action Recognition in Still Images} 
\titlerunning{Loss Guided Activation for Action Recognition in Still Images} 

\author{Lu Liu$^1$, Robby T. Tan$^{1,2}$, Shaodi You$^{3,4}$}
\institute{$^1$ECE Department, National University of Singapore, $^2$Yale-NUS College, Singapore\\
	$^3$DATA61-CSIRO, $^4$Australian National University, Australia\\
	\email{lliu@u.nus.edu, robby.tan@nus.edu.sg, shaodi.you@data61.csiro.au}
}

%

\authorrunning{Lu Liu, Robby T. Tan, Shaodi You} 

%

\maketitle

\begin{abstract}
One significant problem of deep-learning based human action recognition is that it can be easily misled by the presence of irrelevant objects or backgrounds. 
Existing methods commonly address this problem by employing bounding boxes on the target humans as part of the input, in both training and testing stages. This requirement of bounding boxes as part of the input is needed to enable the methods to ignore irrelevant contexts and extract only human features. 
However, we consider this solution is inefficient, since the bounding boxes might not be available.
Hence, instead of using a person bounding box as an input, we introduce a human-mask loss to automatically guide the activations of the feature maps to the target human who is performing the action, and hence suppress the activations of misleading contexts.
We propose a multi-task deep learning method that jointly predicts the human action class and human location heatmap. Extensive experiments demonstrate our approach is more robust compared to the baseline methods under the presence of irrelevant misleading contexts. 
Our method achieves 94.06\% and 40.65\% (in terms of mAP) on Stanford40 and MPII dataset respectively, which are 3.14\% and 12.6\% relative improvements over the best results reported in the literature, and thus set new state-of-the-art results.
Additionally, unlike some existing methods, we eliminate the requirement of using a person bounding box as an input during testing.
\end{abstract}
\section{Introduction}
Action recognition from a single image is generally still challenging. An input image can contain multiple objects and humans, with occlusions, cluttered backgrounds, viewpoint variations, and articulated human poses, making the task of action recognition much more challenging than standard image classification task. Existing methods have exploited cues such as human body pose~\cite{desai2012detecting,yao2012action,girdhar2017AttentionalPoolingAction}, interactive objects~\cite{oquab2014learning,gkioxari2015contextual}, body part appearances~\cite{maji2011action,gkioxari2015actions,zhao2017single}, and multiple instance learning~\cite{mallya2016learning} to handle the aforementioned problems.

\begin{figure}[ht!] 
         \centering 
         \includegraphics[width=1\columnwidth]{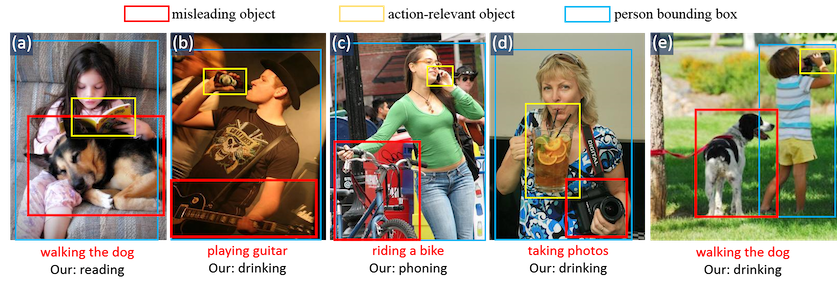}
         \caption{Examples of irrelevant objects that mislead human action predictions. The wrong predictions using previous holistic method~\cite{mallya2016learning} are marked in red. Our proposed method implicitly suppresses the activations of misleading contexts and therefore can correctly identify the actions (marked in black).}
        \label{fig:eg} 
\end{figure}

Particularly for deep-learning based methods, human action recognition can be misled by irrelevant objects or backgrounds in the input image. Mining contextual object cues can be helpful to recognize human actions that involve objects, but can also be unreliable under the presence of misleading contexts. An input image may contain multiple objects, some of which are relevant and discriminative to recognize the action, but some are irrelevant and misleading to recognize the action. Some examples of misleading objects and action-relevant objects are shown in Fig.~\ref{fig:eg}. These irrelevant cues can be salient (e.g. the dog in Fig.~\ref{fig:eg}(a)(e)) and interactive with human (e.g. the dog, guitar, bike, camera in Fig.~\ref{fig:eg}(a-d)), making them even harder to be ignored by recognition algorithms.
Intuitively, human action prediction should focus on the target humans with priority. However, most of the existing methods are significantly driven by training data, which can be biased for non-human objects and backgrounds. Consequently, instead of focusing on humans, the attention of the algorithms can be shifted to irrelevant contextual cues, leading to a wrong prediction.

To address the misleading contexts problem, existing approaches usually use person bounding boxes as input~\cite{maji2011action,hoai122014regularized,oquab2014learning,gupta2015visual,gkioxari2015contextual,yang2016exploit,mallya2016learning,sharma2017expanded,zhao2017single} in both training and testing stages. This is needed to extract features of the target human and then to combine them with contextual features from the whole image. However, we consider that using a person bounding box as an additional input is not effective to exclude irrelevant cues in the image. This form of hard attention does not tackle the underlying problem, because the extracted features of the misleading contexts can have higher response than action-related contexts, due to some probable bias in the dataset for object manipulation type of actions, leading to a wrong prediction.

Additionally, the appearance of objects have much fewer variations and thus higher consistency than the appearance of human body in the training data. As a result, deep neural networks learn richer representations of objects and other contexts than the human body.
The unbalanced feature activations among objects and humans make the existing methods sensitive to the irrelevant misleading contexts, and hence these methods perform poorly under their presence. 
For example, in Fig.~\ref{fig:eg}(a), the presence of a dog makes the action to be misclassified as "walking the dog" by previous holistic methods~\cite{mallya2016learning,girdhar2017AttentionalPoolingAction}.

In this paper, our goal is to divert the activations/attentions of the network towards the target human, and learn its rich deep representations, as well as simultaneously learn compact representations of action-relevant objects and contexts. We propose a multi-task deep learning framework that jointly predicts the human action class and human location heatmap (Fig.~\ref{fig:network}). Instead of adding the person bounding boxes in the input~\cite{mallya2016learning,zhao2017single} or modifying the extracted feature maps by multiplying them with an saliency map~\cite{girdhar2017AttentionalPoolingAction}, we use a novel human-mask loss to automatically guide the activations of the feature maps to the human who is performing the action, and hence suppress the influences of the misleading objects or backgrounds. To our knowledge, it is the first time that we explicitly
show the class activation map can be influenced by a human-mask loss. The practical benefit of this is that we do not need bounding boxes during testing.
Evaluations on two popular and challenging datasets: Stanford40 Action dataset~\cite{yao2011human} and MPII Human Pose dataset~\cite{andriluka20142d} show the effectiveness of our method.
To sum up, our main contributions are three-fold:
\begin{itemize}
\item We propose a new human-mask loss to automatically guide the activation of the network into the human regions to learn rich deep representations of humans. This eliminates the requirement of bounding boxes as part of the input in testing stage.
\item We propose a multi-task deep learning method that jointly predicts the action class and human location heatmap.
\item Our method achieves 94.06\% and 40.65\% (in terms of mean Average Precision, mAP) on Stanford40 and MPII dataset respectively, which are 3.14\% and 12.6\% relative improvements over the best results reported in the literature, and thus set new state-of-the-art results.
\end{itemize}

The rest of the paper is organized as follows. Section~\ref{sec:related_work} reviews the related work of action recognition from still images. Section~\ref{sec:method} describes our approach, which is a multi-task learning framework. Section~\ref{sec:results} shows experimental results and our evaluations quantitatively and qualitatively. Finally, Section~\ref{sec:conclusion} provides a brief summary of our method and some practical future works.

\section{Related work}
\label{sec:related_work}

Compared to action recognition from videos, which highly relies on motion, action recognition from a single image depends on static cues, such as human pose, body parts, and interactive objects. Existing methods can be grouped into three categories: holistic methods, part-based methods, context-based methods. 

{\bf Holistic methods:} Holistic methods extract features from the human in the given bounding box and combine them with contextual features from the whole image to predict human actions~\cite{desai2012detecting,yao2012action,mallya2016learning}. Early works~\cite{desai2012detecting,yao2012action} use a graphical model on the human body pose to infer actions. Recently, Mallya and Lazebnik~\cite{mallya2016learning} propose a simple fusion network that concatenates features extracted from a bounding box with features from the whole image for action prediction. Overall, holistic methods follow the most straightforward strategy and do not involve many pre-processing steps. However, holistic methods can be easily misled by the presence of irrelevant objects or backgrounds. To resolve this problem, our approach introduces a human-mask loss to guide the activations of the network into human regions, and hence suppresses the response of irrelevant contexts.

{\bf Part-based methods:} Part-based approaches detect multiple bounding boxes on various body parts and combine their features with global features to predict actions~\cite{maji2011action,gkioxari2015actions,zhao2017single}. Gkioxari et al.~\cite{gkioxari2015actions} train body part detectors on 'pool5' features in a sliding window manner and combine them with the ground-truth box to train a CNN for action classification. Recently, Zhao et al.~\cite{zhao2017single} incorporate mid-level body part actions (e.g. head: laughing) to infer body actions. However, this method requires an external human pose estimation technique to localize body keypoints and crop out part patches in both training and testing stages. Moreover, the "hard-coded attention" limits the regions to be around the human. Instead of using body parts' patches as input, our approach learns rich representations of humans by using our human-mask loss. 

{\bf Context-based methods:} Contextual algorithms exploit contextual cues, such as interactive objects. CAI~\cite{zhuang2017towards} utilizes language information of the context (i.e.
subject and object) labels, and encodes them into semantic space to learn
context-dependent classifier for visual relationship detection. R*CNN~\cite{gkioxari2015contextual} applies selective search~\cite{uijlings2013selective} to generate object proposals to discover proper interactive objects. However, these proposals are required for both training and testing stages, and the sampling over potential proposals might also be computationally expensive. Moreover, R*CNN uses two hyper-parameters to define the overlap between the person bounding box and the proposal box. Our approach achieves this overlapping by introducing a human-mask loss, which can automatically divert the attention into the most discriminative image regions around the human, in a soft and learnable way.

{\bf Weakly-supervised localization:} All the aforementioned methods require the prior knowledge of the ground-truth bounding boxes in both training and testing stages, making them difficult to scale to real-world applications. There have been a number of recent works exploring weakly-supervised object localization or soft attention~\cite{oquab2014learning,zhang2016action,girdhar2017AttentionalPoolingAction}. Oquab et al.~\cite{oquab2014learning} transfer mid-level image representations obtained from image classification to action recognition. Zhang et al.~\cite{zhang2016action} generate a foreground action mask using a five-step iterative optimization method, then extract features from the action mask for recognition purpose. However, this method suffers from high optimization complexity. Recently, Girdhar and Ramanan~\cite{girdhar2017AttentionalPoolingAction} propose a pooling method that scales the score map with a saliency map. This method potentially assumes that the salient objects are the most useful cues for identifying actions. However, there could be salient but irrelevant objects (see Fig.~\ref{fig:eg}) that can lead to wrong predictions. Our approach implicitly models attention via a number of feature activation maps. We show that it is unnecessary to explicitly model the attention map, but by training the network to predict the human location heatmap. Doing this, we implicitly divert the attention from the misleading contexts to the human regions.

{\bf Multi-task Learning:} Some prior works have shown that jointly learning multiple tasks that relate to each other boosts the individual performances of all the tasks. To name a few, HyperFace~\cite{ranjan2017hyperface} jointly learns face detection, landmarks localization, pose estimation and gender recognition tasks, and improves individual performances. Simonyan and Zisserman~\cite{simonyan2014two} use multi-task learning to decrease over-fitting by jointly training two video datasets. We observe a similar performance boost, where a multi-task learning approach detecting human, as a by-product, improves action classification performance. By jointly predicting the location of the human, the network learns rich representations of the human who is performing the action, and thus achieves better action prediction results.
\section{Our Approach}
\label{sec:method}
The presence of misleading objects or backgrounds can pose a major problem for human action recognition. To address this, existing methods attempt to turn the focus more on the target human. Their strategies to turn the focus on the target human can be categorized into: 
Input modification and feature modification. An example of input modification is Zhao et al.'s method~\cite{zhao2017single}, which crops the region of the given person bounding box to extract the features of the human. Another example is an approach by \cite{gkioxari2015contextual,mallya2016learning}, which uses box coordinates and a Regions Of Interest (ROI) pooling layer~\cite{girshick2015fast} on top of the last convolutional feature maps to extract features on the human. One example of feature modification is to reweight the extracted image feature maps by scaling them with either a human pose heatmap or with a saliency map~\cite{girdhar2017AttentionalPoolingAction}.

However, neither input modification nor feature modification can resolve the problem. Since, for input modification, a person bounding box may still include irrelevant contexts due to the viewpoint, and the close spatial relationship between the human and these contexts. For feature modification, a saliency map produced by a data-driven deep learning model can magnify the effect of misleading contexts rather than suppress them, and hence, can lead to incorrect action predictions. 

\begin{figure}[ht!] 
         \centering 
         \includegraphics[width=1\columnwidth]{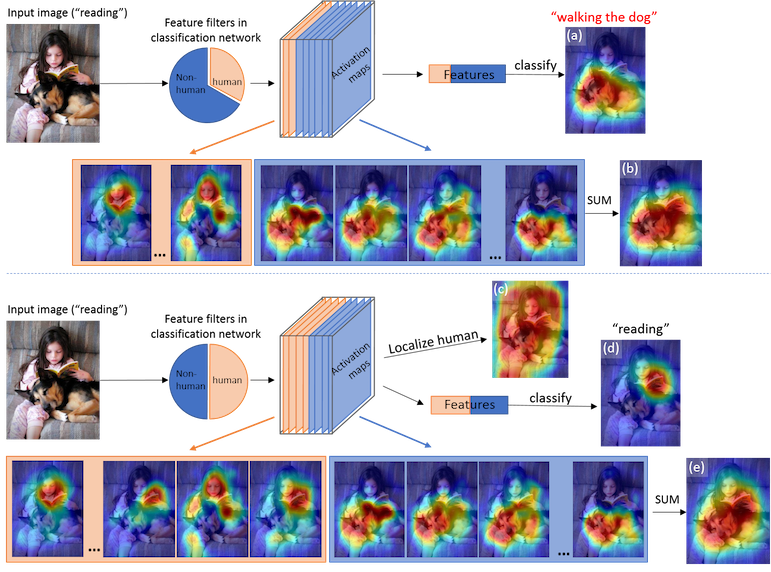}
         \caption{An illustration of a classification network trained without and with our human-mask loss. The multi-task learning framework achieves balanced activations between humans and non-human contexts, which is more robust under the presence of irrelevant misleading contexts.}
        \label{fig:idea} 
\end{figure}

\subsection{Key Idea: Human-mask Loss}
Our key idea is to use a novel human-mask loss to automatically divert the activation of the network into the human regions to learn rich representations of humans, as illustrated in Fig.~\ref{fig:idea}. Under the guidance of human-mask loss, the network is forced to learn more features of humans in order to produce the final human location heatmap, and hence enhance the influence of humans in the final decision. After all, human action recognition must be firstly about human, not the surrounding objects or backgrounds.

As illustrated in Fig.~\ref{fig:idea}, by visualizing the Sum of Activation Maps (SAM) of the network trained with only an action classification loss (Fig.~\ref{fig:idea}(b)), we observe that the final feature maps have much higher activations on salient objects (see the dog), but much lower activations on the human body. This unbalanced activation could probably be the reason why the existing deep learning methods~\cite{oquab2014learning,gkioxari2015contextual,mallya2016learning,girdhar2017AttentionalPoolingAction}  are fragile to misleading contexts. 

We input only the whole image into the network, by training the network to predict the human location heatmap, we encourage the network to learn rich representations about humans (see the highlighted face of the reading girl in Fig.~\ref{fig:idea}(e) compared to (b)). Thus, with the balanced activations on humans and contexts, the network gives the correct action prediction, and the final Predicting Activation Map (PAM) (i.e. the Class Activation Map (CAM~\cite{zhou2016learning}) of the predicted class) shifts attention from the irrelevant objects or backgrounds (e.g. the dog in Fig.~\ref{fig:idea}(a)) to the human's body parts (e.g. the holding hand in Fig.~\ref{fig:idea}(d)), as well as the action-related interactive objects (e.g. the book in Fig.~\ref{fig:idea}(d)) around that human.

\subsection{Network Architecture}
\begin{figure}[t] 
         \centering 
         \includegraphics[width=1\columnwidth]{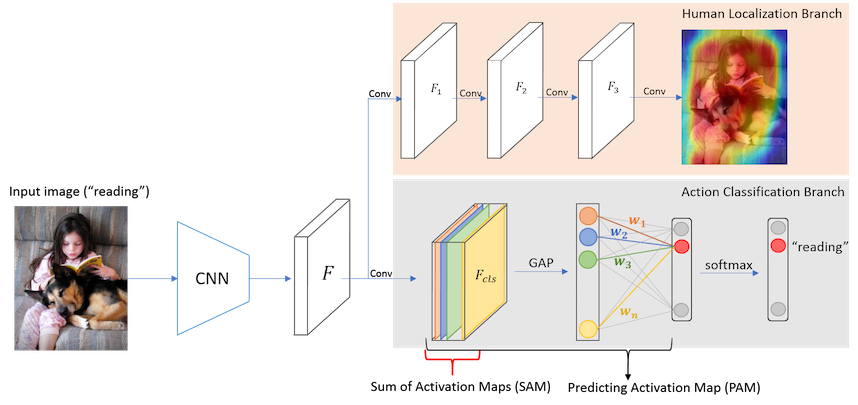}
         \caption{Network architecture of the proposed human-mask loss guided activation network. During training, given an input image, the network is trained to predict an action class guided by the ground-truth action label, and a human heatmap guided by the binary human mask. The groundtruth human mask images for the training data are generated using the person bounding box information given in the dataset. During testing, given an input image, the network jointly predicts the action class and human location heatmap.}
        \label{fig:network} 
\end{figure}

The architecture of our proposed network is shown in Fig.~\ref{fig:network}. There are two branches in the network: Action classification branch that predicts the action class, and human localization branch that produces the human location heatmap. Given an input image, we first use a CNN (Inception-ResNet-v2~\cite{szegedy2017inception}) to extract feature maps \addtocounter{footnote}{+0}\footnotemark  from the last convolutional layer. By jointly predicting the human location heatmap, the network is forced to learn rich representations of humans, and hence suppresses the influences of irrelevant misleading contexts.
\footnotetext{This backbone feature is shared by both two branches.}

{\bf Action Classification Branch}: On top of the backbone features $F$, we use a convolutional layer to further reduce the number of channels to extract compact features $F_{cls}$ for classification. Since this convoluational layer is only trained using classification loss, it provides the classification task with more flexibility and capacity. Then, we perform global average pooling (GAP) on the feature maps $F_{cls}$ to obtain a feature vector $V$, and use it to train a softmax classifier to predict the action class. We use only one fully-connected (FC) layer for predicting action labels, so that the weights of the FC layer can be projected back on to the convolutional feature maps $F_{cls}$, indicating image regions that have been used by the network to recognize that action class. 

{\bf Human Localization Branch:} Our goal is to divert the activations of the feature maps into the human regions to learn rich representations of the target human and the surrounding interactive objects. To accomplish this, we add the human localization branch to create a human heatmap guided by the binary human mask $M^{gt}$ (we use $gt$ to denote ground-truth). In the mask, $M^{gt}(i) = 1$ means the pixel $i$ is inside the person bounding box\footnote{The person bounding box coordinates are given by the dataset.}, otherwise it belongs to the background regions. Note that, we only generate this groundtruth human mask for training data. Based on the backbone feature maps $F$, we further apply four convolutional layers to generate a 2D human location heatmap $M^*$. To obtain a mask with a proper spatial resolution, these convolutional layers preserve the spatial dimension and only reduce the number of channels gradually. Finally, we compute the L2-norm distance between the output map $M^*$ and the ground-truth mask $M^{gt}$ and back-propagate the error.

{\bf Loss Function:} We use cross-entropy loss for action classification task, and the L2-norm distance between the predicted human-mask $M^*$ and the ground-truth human mask $M^{gt}$ as the loss function for human localization task (Eq.~\ref{eq:score}). We combine the two losses with equal weights $L = L_{cls} + L_{mask}$, where:

\begin{equation} \label{eq:score}
L_{cls}  = - \log (\frac{\text{exp}(S_{c^{gt}})}{\sum_c \text{exp}(S_c)}); \quad \quad   
L_{mask} = || M^{gt} - M^*||_2^2,
\end{equation}
where $S_c$ is the score before softmax of class $c$, and $c^{gt}$ is the ground-truth class.

\subsection{Loss-guided Activation}
We summarize all channels of the final activation map $F_{cls}$ to obtain a 2D map, denoted as SAM (Sum of Activation Maps). By visualizing SAM, we are able to evaluate the distribution of the feature kernels learnt by the network trained with and without our human-mask loss. To investigate based on which image regions that the CNN is making its decision, we further compute the weighted sum of the activation maps at the predicted class $c^*$ (i.e. CAM at the predicted class), denoted as PAM (Predicting Activation Map). Here are the definitions of SAM and PAM, respectively:
\begin{equation} \label{eq:sam}
\text{SAM}(i,j) = \sum_k F^k_{cls} (i,j);  \quad \quad \quad
\text{PAM}(i,j) = \sum_k w^k_{c^*} F^k_{cls}(i,j),
\end{equation}
where $F^k_{cls}$ is the $k$th channel of the final activation map $F_{cls}$, $(i,j)$ is the spatial location, $c^*$ is the predicted action class, and $w^k_{c^*}$ is the learnt weight of the $k$th feature for predicted class $c^*$.

\section{Experiments}
\label{sec:results}
We use two challenging action datasets: (1) Stanford40 Action Dataset~\cite{yao2011human} consisting of 9532 images of people performing 40 actions. The dataset is split into training and test sets with 4000 and 5532 images each. (2) MPII Human Pose Dataset~\cite{andriluka20142d} containing 20,916 images classified into one of the 393 action classes. It is split into training, validation (from authors of~\cite{gkioxari2015contextual}) and test sets, with 8219, 6988 and 5709 images each. The final test mAP results are obtained by emailing our results to authors of~\cite{andriluka20142d}. The annotations do not include a ground-truth bounding box explicitly, but provide the location of 16 human body keypoints. This information is used to generate human-mask images for the training data. Among all coordinates of body joints, the min and max coordinates are picked to composite a tight box covering the human body joints. Then we expand the box by 50\% to cover the whole body, and generate human-mask images for training.

To obtain the final activation maps of resolution $14\times14$, the test images are resized to $448 \times 448$ and inputted to the network. We train two backbone CNNs: ResNet~\cite{he2016deep} and Inception-ResNet-v2~\cite{szegedy2017inception} initialized with ImageNet~\cite{deng2009imagenet} weights. On top of the backbone feature maps, (i) In action classification branch, we use one convolutional layer with 1024 kernels ($3 \times 3$ kenel size, stride 1) and ReLu nonlinearity to obtain the final feature maps $F_{cls}$ ($1024\times14\times14$). This $F_{cls}$ is then global average pooled to a feature vector for training the softmax classifier; (ii) In human localization branch, we apply four Conv-ReLu layers (all $3 \times 3$ kernel size, stride 1) to gradually reduce the channel numbers ($512 \rightarrow 64 \rightarrow 32 \rightarrow 1$) to generate the final human location heatmap. The learning rate is set to be $10^{-5}$, and batch size is 12. Three kinds of data augmentations are employed: horizontal flipping, random rotation (range of 0-10 degrees), and random zoom (0.9-1.1). 

\subsection{Comparisons with Existing Methods}
\begin{figure}[ht!] 
         \centering 
         \includegraphics[width=1\columnwidth]{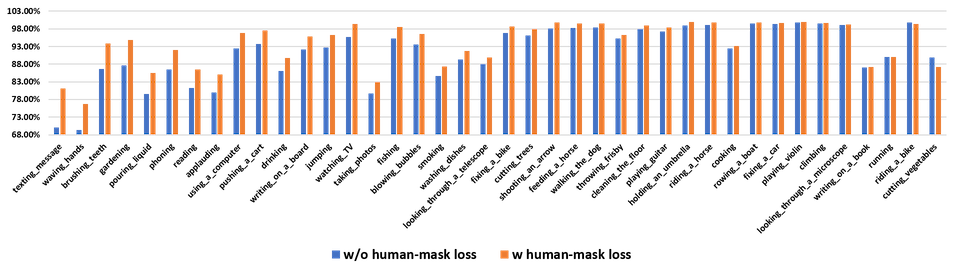}
         \caption{AP (\%) comparison between a network (Inception-ResNet-v2 based) trained with and without our human-mask loss on Stanford40 dataset. The results of all actions are shown in descending order of their absolute AP improvements. The mean AP improvement across all actions is 2.64\%.}
        \label{fig:map_stanford} 
\end{figure}

{\bf Stanford40 Action Dataset.} Table~\ref{tab:stanford} shows the results on Stanford40 dataset~\cite{yao2011human}. Using Inception-ResNet-v2 as backbone CNN, our method achieves a mAP of 94.06\% on Stanford40 test set, which is the state-of-the-art. Performance varies from 76.7\% for "waving hands" to 100\% for "playing violin". For all the 40 categories, the improvement of using our human-mask loss comes from two sources: (1) Test samples that contain irrelevant misleading objects and backgrounds; (2) Confusing action pairs such as "waving hands" and "applauding". Fig.~\ref{fig:map_stanford} shows the AP performance per action on the test set. In comparison with previous best approach PAN~\cite{zhao2017single} (mAP of 91.2\%), which uses bounding boxes in the input image, our method’s performance is comparable (mAP of 91.1\%). In fact, PAN uses body part bounding boxes (in addition to the person bounding boxes) and additional body part action annotations, thus ours uses less information. The benefit of our method compared to PAN is that we do not need the bounding boxes in the testing, which in terms of practicality is a significant improvement.

{\bf Effectiveness of our human-mask loss.} We train action classification network with/without our human-mask loss to compare the effectiveness of our introduced human-mask loss. Our human-mask loss improves the mAP by 2.3\% and 2.64\% for both ResNet50 and Inception-ResNet-v2 based network respectively. Jointly predicting human location heatmap significantly boosts action classification performance. Fig.~\ref{fig:map_stanford} shows the AP comparison between a network trained with and without our human-mask loss. Our method significantly improves mAP on the top confusing pair "waving hands" and "applauding" by 7.33\% and 5.06\% respectively. It also obtains large gains on object manipulation type of actions, such as "texting message" (+11.09\%), "brushing teeth" (+7.31\%), "pouring liquid" (+5.98\%), "phoning" (+5.59\%). There is an accuracy drop for "cutting vegetables". The misclassification happens because the knife is lying on the table and the hand is holding the vegetables.

\begin{table}[!t]
\centering
\caption {mean Average Precision (mAP) on Stanford40 dataset}
\label{tab:stanford}
\begin{tabular}{lc}
\toprule
{\bf Methods} &  {\bf mAP(\%)} \\ \midrule
Action Mask~\cite{zhang2016action}&  82.64\\
ResNet50\footnotemark~\cite{he2016deep} & 87.23 \\
Inception-ResNet-v2\footnotemark[\value{footnote}]~\cite{szegedy2017inception} & 90.38\\
VGG-16, R*CNN~\cite{gkioxari2015contextual} &  90.90\\
ResNet50, Part Action Network~\cite{zhao2017single} & 91.20 \\ \midrule
Ours - ResNet50, w/o human-mask loss & 88.80 \\
Ours - ResNet50, w human-mask loss &  91.10\\
Ours - Inception-ResNet-v2, w/o human-mask loss \quad \quad \quad &  91.42\\
{\bf Ours - Inception-ResNet-v2, w human-mask loss} &  \textbf{94.06}\\
\bottomrule
\end{tabular}
\end{table}
\footnotetext{A standard classification network, trained with the same experimental configurations as ours, but without adding one convolutional layer on top of the backbone CNN.}

{\bf MPII Dataset.} Table~\ref{tab:mpii} shows the comparison on MPII test and validation sets. We use the validation set shared by the authors of~\cite{gkioxari2015contextual} to compare with~\cite{gkioxari2015contextual,girdhar2017AttentionalPoolingAction}. Performance on the test set is obtained by submitting our prediction scores to authors of~\cite{andriluka20142d}. The previous best approach is Attn.Pool~\cite{girdhar2017AttentionalPoolingAction}, which achieves a mAP of 36.1\% on test set. Using Inception-ResNet-v2 as backbone CNN with our human-mask loss, our method achieves a mAP of 40.65\% on MPII test set, surpassing previous benchmark by 12.6\% (relative improvement). 

{\bf Effectiveness of our human-mask loss.} Our human-mask loss improves the mAP by 1.21\% and 1.90\% for both ResNet101 and Inception-ResNet-v2 based network respectively on validation set. For all 393 categories, we observe that the top improved actions are those whose critical cues are about humans rather than contexts, which may contain irrelevant objects and cluttered backgrounds. For example, our human-mask loss significantly improves "sitting, in class, general, including note-taking or class discussion" by 40.1\%, "woodwind, sitting" by 36.4\%, "laughing, sitting" by 25.58\% on validation set. 

\begin{table}[!t]
\centering
\caption {mean Average Precision (mAP) on MPII dataset}
\label{tab:mpii}
\begin{tabular}{lcc}
\toprule
{\bf Methods} &  {\bf Val mAP(\%)} &  {\bf Test mAP(\%)}\\ \midrule
Dense Trajectories+Pose~\cite{pishchulin2014fine} &-&5.5  \\
VGG-16, Scene-RCNN~\cite{gkioxari2015contextual} &16.5&-  \\
VGG-16, R*CNN\footnotemark~\cite{gkioxari2015contextual} &21.7&26.7\\
VGG-16, Fusion~\cite{mallya2016learning} & - &32.3 \\
Inception-v2, Attn.Pool~\cite{girdhar2017AttentionalPoolingAction} & 24.3 & - \\
ResNet101, Attn.Pool~\cite{girdhar2017AttentionalPoolingAction} & 30.3 & 36.0\\
ResNet101, Attn.Pool+Pose~\cite{girdhar2017AttentionalPoolingAction} & 30.6 & 36.1\\ \midrule
Ours - ResNet101, w/o human-mask loss & 30.77  & - \\
Ours - ResNet101, w human-mask loss & 31.98  &-\\
Ours - Inception-ResNet-v2, w/o human-mask loss &  32.38 &-\\
{\bf Ours - Inception-ResNet-v2, w human-mask loss} & \textbf{34.28} &\textbf{40.65}\\
\bottomrule
\end{tabular}
\end{table}
\footnotetext{R*CNN reports the test AP of 1.1\% for both "cooking or food preparation" and "video exercise workout" actions, while our method achieves 25.64\%, and 11.11\% on the two action classes respectively.}

\subsection{Visualization of Activation Maps}
\begin{figure}[ht!] 
         \centering 
         \includegraphics[width=1\columnwidth]{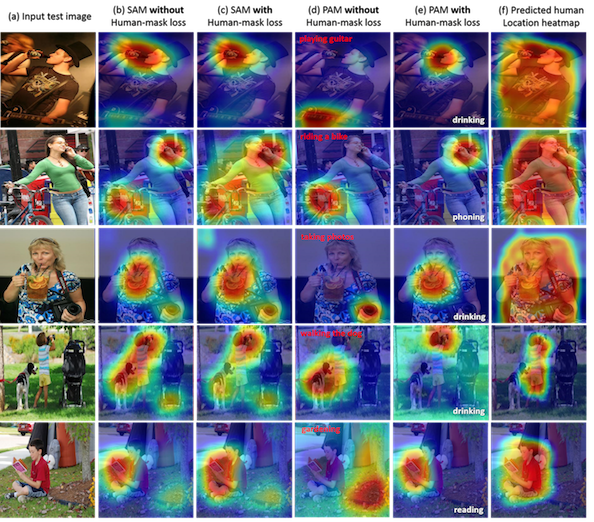}
         \caption{Examples of the SAM and PAM obtained from the network trained with/without human-mask loss. Wrong predictions are marked in red, and correct ones (ours) are marked in white. Using our human-mask loss, the final predicting attention shifts from the misleading objects (e.g. guitar, bike, camera, dog) or backgrounds (e.g. garden) to the human regions.}
        \label{fig:shift} 
\end{figure}

We visualize the activation maps of the network (Inception-ResNet-v2 based) trained with/without our human-mask loss. Fig.~\ref{fig:shift} shows the shifted attention on SAM and PAM using our human-mask loss. Note for a fair comparison between a classification network trained with and without a human-mask loss, we use min-max normalization to normalize each channel of $F_{cls}$ to [0,1] before summation. Given an input image as shown in Fig.~\ref{fig:shift}(a), the network jointly predicts the action class and human location heatmap as shown in Fig.~\ref{fig:shift}(f). By comparing SAM trained with and without a human-mask loss in Fig.~\ref{fig:shift}(b)(c), we observe that our human-mask loss successfully drives more activations into the human regions, such as the boy besides the dog, and the human carrying the guitar. Therefore, the final PAM (Fig.~\ref{fig:shift}(d)(e)) shifts the attention from the misleading objects (e.g. guitar, bike, camera, dog) or backgrounds (e.g. garden) into the human regions. Our proposed human-mask loss guides the network to focus more into the discriminative image regions where humans and the non-human contexts have balanced contributions for predicting actions. 

\begin{figure}[ht!] 
         \centering 
         \includegraphics[width=1\columnwidth]{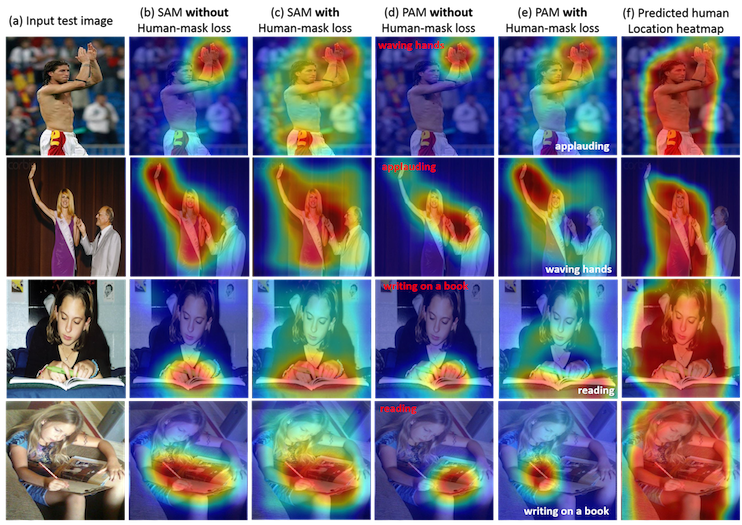}
         \caption{Examples of two confusing action pairs. Wrong predictions are marked in red, and correct ones (ours) are marked in white. Surprisingly, by attending on humans, the network captures more discriminative body pose features, which helps distinguish between "applauding" and "waving hands", as well as the specific way of interaction (holding a pen or writing with a pen), which helps distinguish between "reading" and "writing on a book".}
        \label{fig:conf_pairs} 
\end{figure}

\begin{figure}[ht!] 
         \centering 
         \includegraphics[width=1\columnwidth]{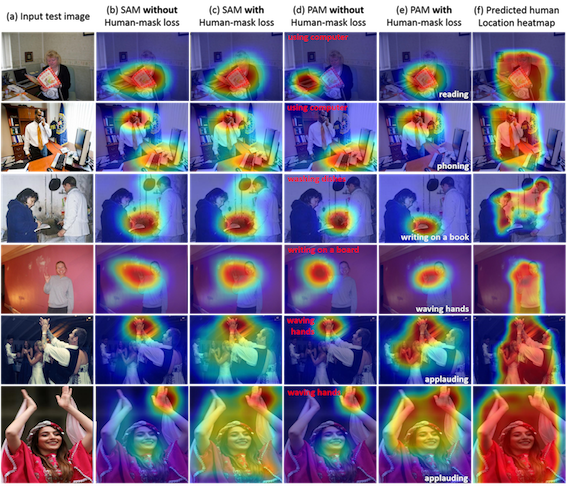}
         \caption{Further examples on Stanford40 test set. First four rows show the cases of misleading contexts, and last two rows show a case of confusing action pair. Wrong predictions are marked in red, and correct ones (ours) are in white.}
        \label{fig:sup_stanford} 
\end{figure}

\begin{figure}[ht!] 
         \centering 
         \includegraphics[width=1\columnwidth]{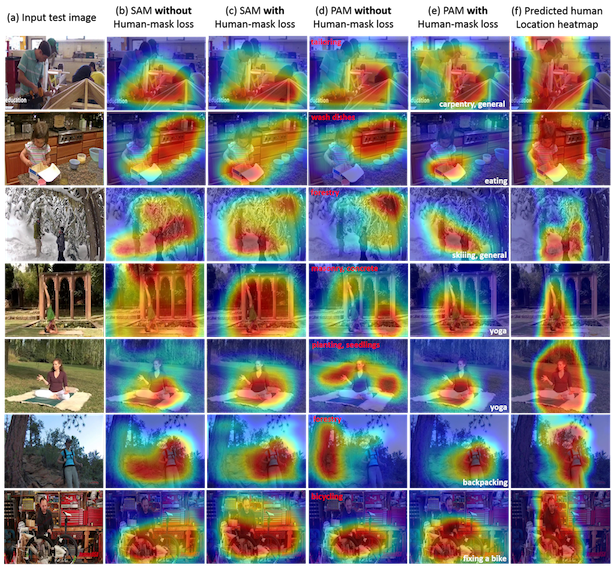}
         \caption{Further examples on MPII validation set. Wrong predictions are marked in red, and correct ones (ours) are marked in white. Our human-mask loss guides the activations from the misleading objects (e.g. sink) or backgrounds (e.g. forest) to the humans and action-relevant interactive objects (if any).}
        \label{fig:sup_mpii} 
\end{figure}

Additionally, we observe some corrections using human-mask loss benefit from learning better representations of humans. Fig.~\ref{fig:conf_pairs} shows examples of two confusing pairs of "applauding" vs. "waving hands", and "reading" vs. "writing on a book". For instance, Surprisingly, by attending on humans, the network captures more discriminative body pose features, which helps distinguish between "applauding" and "waving hands". The key to distinguish between "applauding" and "waving hands" is the pose of the upper body. Usually, "waving hands" requires one hand, while "applauding" requires two hands. Under the guidance of our human-mask loss, the network is able to capture the overall pose of the human's upper body rather than purely focusing on the local hand regions. Further examples can be found in Fig.~\ref{fig:sup_stanford} and Fig.~\ref{fig:sup_mpii}.
%

\subsection{Comparison and Discussion}

\begin{figure}[ht!] 
         \centering 
         \includegraphics[width=1\columnwidth]{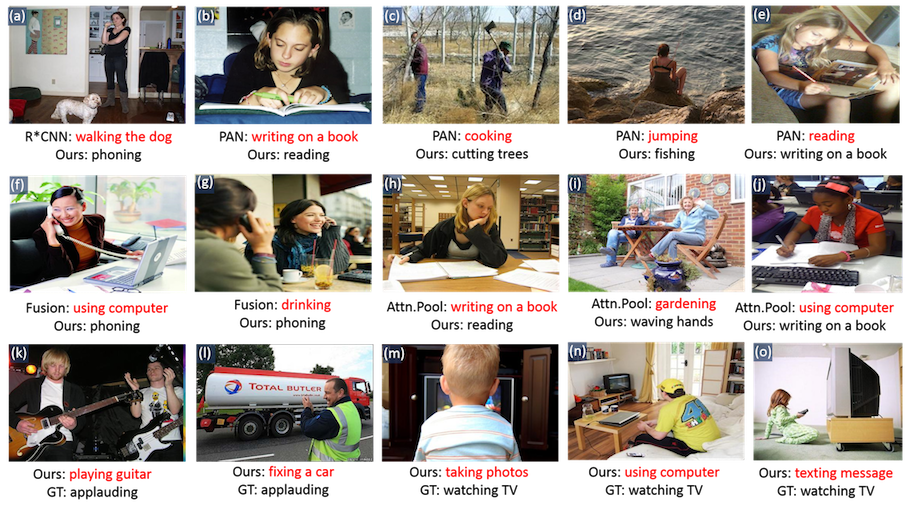}
         \caption{Predictions obtained by our method and existing methods on Stanford40 test set. The correct predictions are marked in black, and wrong ones are in red. Results show that our method is more robust than existing methods under the presence of irrelevant misleading contexts (see first two rows). Our approach also has certain limitations when misleading objects are too dominant, or action-relevant objects are largely occluded, or have no direct interaction with the human in the presence of multiple objects in the image (see third row).}
        \label{fig:compare} 
\end{figure}

We show some predictions obtained by our method and existing methods in Fig.~\ref{fig:compare}. R*CNN~\cite{gkioxari2015contextual} can misclassify an action when the misleading objects are selected as its secondary box with highest response score (Fig.~\ref{fig:compare}(a)). PAN~\cite{zhao2017single} focuses on local body parts and can misclassify when body parts are occluded (Fig.~\ref{fig:compare}(b-e)). Fusion~\cite{mallya2016learning} can make a wrong prediction when the misleading objects are inside the person bounding box (Fig.~\ref{fig:compare}(f)(g)). Attn.Pool~\cite{girdhar2017AttentionalPoolingAction} can magnify the response of misleading contexts, leading to a wrong prediction (Fig.~\ref{fig:compare}(h-j)). Compared to the aforementioned methods, our method applies human-mask loss and successfully diverts the activations of the network to human, and hence gives correct action predictions, as shown in (Fig.~\ref{fig:compare}(a-j)).

In the last row of Fig.~\ref{fig:compare}(k-o), we show some misclassified samples by our method. There are mainly three reasons: (1) Misleading objects are too dominant (i.e. occupy a larger portion of the image than the human does) to be ignored (see the big car in front of the applauding man in Fig.~\ref{fig:compare}(k)). (2) Action-relevant objects are largely occluded (the brush and TV in Fig.~\ref{fig:compare}(l)(m)). (3) Indirect interaction betwen human and action-relevant objects in the presence of multiple objects (Fig.~\ref{fig:compare}(n)(o)). Our human-mask loss implicitly increases the activations of the objects that are close to the target human. We believe that by explicitly detecting the interactive objects using human-object interaction models such as~\cite{gkioxari2017detecting}, our method can perform even better. We leave this for our future work.

\section{Conclusion}
\label{sec:conclusion}

In this paper, we propose a multi-task learning method to solve the problem of irrelevant misleading contexts for action recognition in still images. Our goal is to divert the activations of the network to focus on humans, and hence the activations of the misleading objects or backgrounds can be suppressed. We introduce a novel human-mask loss to automatically guide the activations of the feature maps to the target human. We propose a multi-task deep learning method that jointly predicts the human action class and human location heatmap. Our method achieves state-of-the-art results: 94.06\% on Stanford40 and 40.65\% on MPII dataset, surpassing the previous benchmarks. Additionally, we eliminate the requirement of using a person bounding box as an input in the testing stage. Future work involves combining human-object interaction technique to better exploit action-relevant contexts in the given images.

\section{Acknowledgement }
This  research  is  supported  by  the  National  Research  Foundation, Prime  Ministers  Office,  Singapore  under  its  Strategic Capability  Research  Centres  Funding Initiative. R.T. Tan's work is supported in part by Yale-NUS College Start-Up Grant. Lu Liu is supported by Yale-NUS College PhD Scholarship.


%
%
%
\clearpage
\bibliographystyle{splncs04}
\bibliography{mybib}

\begin{thebibliography}{10}
\providecommand{\url}[1]{\texttt{#1}}
\providecommand{\urlprefix}{URL }
\providecommand{\doi}[1]{https://doi.org/#1}

\bibitem{andriluka20142d}
Andriluka, M., Pishchulin, L., Gehler, P., Schiele, B.: 2d human pose
  estimation: New benchmark and state of the art analysis. In: Proceedings of
  the IEEE Conference on computer Vision and Pattern Recognition. pp.
  3686--3693 (2014)

\bibitem{deng2009imagenet}
Deng, J., Dong, W., Socher, R., Li, L.J., Li, K., Fei-Fei, L.: Imagenet: A
  large-scale hierarchical image database. In: Computer Vision and Pattern
  Recognition, 2009. CVPR 2009. IEEE Conference on. pp. 248--255. Ieee (2009)

\bibitem{desai2012detecting}
Desai, C., Ramanan, D.: Detecting actions, poses, and objects with relational
  phraselets. In: European Conference on Computer Vision. pp. 158--172.
  Springer (2012)

\bibitem{girdhar2017AttentionalPoolingAction}
Girdhar, R., Ramanan, D.: Attentional pooling for action recognition. In: NIPS
  (2017)

\bibitem{girshick2015fast}
Girshick, R.: Fast r-cnn. In: Computer Vision (ICCV), 2015 IEEE International
  Conference on. pp. 1440--1448. IEEE (2015)

\bibitem{gkioxari2017detecting}
Gkioxari, G., Girshick, R., Doll{\'a}r, P., He, K.: Detecting and recognizing
  human-object interactions. arXiv preprint arXiv:1704.07333  (2017)

\bibitem{gkioxari2015actions}
Gkioxari, G., Girshick, R., Malik, J.: Actions and attributes from wholes and
  parts. In: Proceedings of the IEEE International Conference on Computer
  Vision. pp. 2470--2478 (2015)

\bibitem{gkioxari2015contextual}
Gkioxari, G., Girshick, R., Malik, J.: Contextual action recognition with r*
  cnn. In: Proceedings of the IEEE international conference on computer vision.
  pp. 1080--1088 (2015)

\bibitem{gupta2015visual}
Gupta, S., Malik, J.: Visual semantic role labeling. arXiv preprint
  arXiv:1505.04474  (2015)

\bibitem{he2016deep}
He, K., Zhang, X., Ren, S., Sun, J.: Deep residual learning for image
  recognition. In: Proceedings of the IEEE conference on computer vision and
  pattern recognition. pp. 770--778 (2016)

\bibitem{hoai122014regularized}
Hoai, M.: Regularized max pooling for image categorization. In: Proceedings of
  the British Machine Vision Conference. BMVA Press (2014)

\bibitem{maji2011action}
Maji, S., Bourdev, L., Malik, J.: Action recognition from a distributed
  representation of pose and appearance. In: Computer Vision and Pattern
  Recognition (CVPR), 2011 IEEE Conference on. pp. 3177--3184. IEEE (2011)

\bibitem{mallya2016learning}
Mallya, A., Lazebnik, S.: Learning models for actions and person-object
  interactions with transfer to question answering. In: European Conference on
  Computer Vision. pp. 414--428. Springer (2016)

\bibitem{oquab2014learning}
Oquab, M., Bottou, L., Laptev, I., Sivic, J.: Learning and transferring
  mid-level image representations using convolutional neural networks. In:
  Computer Vision and Pattern Recognition (CVPR), 2014 IEEE Conference on. pp.
  1717--1724. IEEE (2014)

\bibitem{pishchulin2014fine}
Pishchulin, L., Andriluka, M., Schiele, B.: Fine-grained activity recognition
  with holistic and pose based features. In: German Conference on Pattern
  Recognition. pp. 678--689. Springer (2014)

\bibitem{ranjan2017hyperface}
Ranjan, R., Patel, V.M., Chellappa, R.: Hyperface: A deep multi-task learning
  framework for face detection, landmark localization, pose estimation, and
  gender recognition. IEEE Transactions on Pattern Analysis and Machine
  Intelligence  (2017)

\bibitem{sharma2017expanded}
Sharma, G., Jurie, F., Schmid, C.: Expanded parts model for semantic
  description of humans in still images. IEEE transactions on pattern analysis
  and machine intelligence  \textbf{39}(1),  87--101 (2017)

\bibitem{simonyan2014two}
Simonyan, K., Zisserman, A.: Two-stream convolutional networks for action
  recognition in videos. In: Advances in neural information processing systems.
  pp. 568--576 (2014)

\bibitem{szegedy2017inception}
Szegedy, C., Ioffe, S., Vanhoucke, V., Alemi, A.A.: Inception-v4,
  inception-resnet and the impact of residual connections on learning. In:
  AAAI. vol.~4, p.~12 (2017)

\bibitem{uijlings2013selective}
Uijlings, J.R., Van De~Sande, K.E., Gevers, T., Smeulders, A.W.: Selective
  search for object recognition. International journal of computer vision
  \textbf{104}(2),  154--171 (2013)

\bibitem{yang2016exploit}
Yang, H., Tianyi~Zhou, J., Zhang, Y., Gao, B.B., Wu, J., Cai, J.: Exploit
  bounding box annotations for multi-label object recognition. In: Proceedings
  of the IEEE Conference on Computer Vision and Pattern Recognition. pp.
  280--288 (2016)

\bibitem{yao2012action}
Yao, B., Fei-Fei, L.: Action recognition with exemplar based 2.5 d graph
  matching. In: European Conference on Computer Vision. pp. 173--186. Springer
  (2012)

\bibitem{yao2011human}
Yao, B., Jiang, X., Khosla, A., Lin, A.L., Guibas, L., Fei-Fei, L.: Human
  action recognition by learning bases of action attributes and parts. In:
  Computer Vision (ICCV), 2011 IEEE International Conference on. pp.
  1331--1338. IEEE (2011)

\bibitem{zhang2016action}
Zhang, Y., Cheng, L., Wu, J., Cai, J., Do, M.N., Lu, J.: Action recognition in
  still images with minimum annotation efforts. IEEE Transactions on Image
  Processing  \textbf{25}(11),  5479--5490 (2016)

\bibitem{zhao2017single}
Zhao, Z., Ma, H., You, S.: Single image action recognition using semantic body
  part actions. In: 2017 IEEE International Conference on Computer Vision
  (ICCV), Venice. pp. 3411--3419 (2017)

\bibitem{zhou2016learning}
Zhou, B., Khosla, A., Lapedriza, A., Oliva, A., Torralba, A.: Learning deep
  features for discriminative localization. In: Computer Vision and Pattern
  Recognition (CVPR), 2016 IEEE Conference on. pp. 2921--2929. IEEE (2016)

\bibitem{zhuang2017towards}
Zhuang, B., Liu, L., Shen, C., Reid, I.: Towards context-aware interaction
  recognition for visual relationship detection. In: Computer Vision (ICCV),
  2017 IEEE International Conference on. pp. 589--598. IEEE (2017)

\end{thebibliography}

\end{document}